# Contextual Mood Analysis with Knowledge Graph Representation for Hindi Song Lyrics in Devanagari Script

Makarand velankar[1], Rachita Kotian[1], Parag Kulkarni[2]

[1]Cummins College of Engineering for women, Karvenagar, Pune, India.

[2] CEO and Chief Scientist at Kvinna Limited, Pune, Maharashtra, India.

Corresponding Author: makarand.velankar@cumminscollege.in

## Abstract

Lyrics play a significant role in conveying the song's mood and are information to understand and interpret music communication. Conventional natural language processing approaches use translation of the Hindi text into English for analysis. This approach is not suitable for lyrics as it is likely to lose the inherent intended contextual meaning. Thus, the need was identified to develop a system for Devanagari text analysis. The dataset of 300 song lyrics with equal distribution in five different moods is used for the experimentation. The proposed system performs contextual mood analysis of Hindi song lyrics in Devanagari text format. The contextual analysis is stored as a knowledge base, updated using an incremental learning approach with new data. Contextual knowledge graph with moods and associated important contextual terms provides the graphical representation of the lyric dataset used. The testing results show 64% accuracy for the mood prediction. This work can be easily extended to applications related to Hindi literary work such as summarization, indexing, contextual retrieval, context-based classification and grouping of documents.

## Introduction

Music has always been an intricate part of an individual's life. Music powerfully evokes feelings. Choosing an excellent song to set the mood for any occasion has become an essential demand from music listeners. Many people listen to songs in local languages, and analyzing lyrics in local languages is a herculean task. Careful analysis of the lyrics can give us the possible directions related to its sentiment. Lyrics provide high-level information about a song. The only barrier in understanding the lyrics is the language or the style in which it is written. A person with a basic understanding of the language may find it challenging to comprehend the meaning of a song. Contextual analysis of the song can be done to seek out the probable category to describe the mood. It is cumbersome to classify the songs of a language one might not know thoroughly but enjoy listening to them. It will ease if it is automated for the tech-savvy and music-loving generation of today.

Popular or Bollywood songs are enjoyed by the majority of people worldwide. Hindi dataset thus comes as a natural choice to work with Bollywood songs. However, people willing to work on a problem related to the Devanagari script find it challenging to get hold of a suitable data set. As per the survey, it was found out that not much work has been done with the Hindi lyrics in natural language programming compared with English lyrics. However, there are standard steps available for processing data and conducting the contextual analysis. The challenge is to apply those steps over the Devanagari script.

Work on the Hindi language has gained attention recently from different renowned research institutes [1]. There is a good amount of research on different tools and repositories for natural language processing. Working with the Devanagari script provides an opportunity to contribute to this ongoing recent research initiative. Some tools available, like Hindi WordNet API s, are explored to perform the task [2, 3, 4]. It was observed that contextual analysis is not available for the Hindi language. The proposed work aims to contribute to the field of sentiment context analysis for the Hindi language. It can be further extended to help people understand Hindi literature, which sometimes becomes difficult even if they are familiar with the language.

As there was no priorly available dataset for Hindi lyrics in Devanagari script as per our findings, the dataset of about 300 Hindi lyrics in Devanagari script was created for the initial experimentation [5]. This dataset is used for training and testing the algorithm. The dataset included five mood classes associated with the lyrics. The initial knowledge base set in words associated with the sentiments or mood was required to train the model. This knowledge base acted as a repository for words tagged with the probability of the sentiments they may depict. The probability of a word is computed considering the occurrence of the word in each sentiment class lyrics. This knowledge base is used to train the model with a set of lyrics. The classification model is further tested for a new set of lyrics to check the accuracy of the build model. The sentiment with a higher probability is predicted based on the words in the new song lyrics.

The proposed model uses an incremental learning approach and improvises the knowledge base with new songs, thus improving the accuracy. Improvisation of the knowledge base updates the probability weights with words in the song associated with the specific sentiment. The best way to portray the results is by visualizing the data. Creating a knowledge graph [6] is the best choice for such data visualization, as it would, in return, pave the way to a more analytical understanding of the Hindi lyrics. Knowledge graphs even act as a knowledge base for already categorized songs.

This approach helps to reduce the time and computing power used by the code to find results for already known and classified songs. However, one may debate that those Hindi songs can be translated for the people unacquainted with the language and thus question the importance of work. Though these translations translate the Hindi text for the intended audience [7], the actual context or sentiment of the text may be lost in the process. In order to bridge this gap, an intervention is needed. Such sentiment analysis techniques can enhance the experience of the masses trying to enjoy a good song.

Existing systems have implemented the mood analysis for the English song lyrics [8, 9, 10]. Also, many of the available systems have attempted sentiment analysis for Hindi text data, such as analysis on movie reviews [11], documents [12], Twitter [13]. However, till now, little work is being implemented for contextual mood analysis for Hindi song lyrics. Therefore, the proposed experimentation aims with the contextual mood analysis of Hindi song lyrics.

The main contribution of this work can be summarized as

● Devanagari script lyrics dataset

● Algorithm for context identification

● Mood based classification

● Knowledge graph representation

The paper is organized in the following manner. First, the literature survey covers text data analysis approaches with focus on lyrics text. Then, the methodology used covers the steps used for contextual analysis and use of natural language processing (NLP) techniques. Our approach for contextual analysis is explained in subsequent sections. Classification results and incremental learning for knowledge base updating provides the use of machine learning. Next, knowledge graph representation shows visualization for the mood classes with contextual terms identified by the algorithm. Finally, the conclusion and discussion section provide the outcome of the experiments with possible future directions.

## Literature Survey

"A contextual analysis is text analysis to assess the context of historical, cultural setting and also in terms of its textuality or the qualities that characterize the text. It is a systematic study of social, political, economic, philosophical and religious conditions that were in place at the time and place when the text was created" as per [14]. Contextual analysis has many interpretations, and the purpose defines the context to be analyzed, such as document analysis [15], sentiment analysis [16, 17], speech recognition [18]. Contextual analysis means breaking down data and analyzing it to draw the theme [19]. This method is used to find the main idea, the sentiment, or make an informed decision or recommendation [20]. Typically, the steps followed in the contextual analysis are data accumulation, processing, analysis, visualization and conclusions. Various machine learning approaches are explored by researchers for context-based applications such as unsupervised [21], cooperative learning [22] and ensemble learning [23]. Machine learning approaches for music analytics [24] focus on using features for various commercial applications. Contextual analysis for lyric text is a challenging research domain considering the intricacies and interpretations of musical language.

Devanagari script text analysis has been attempted by researchers for different applications to process Hindi text data. Natural language toolkit (NLTK) is a generic platform to process the data of various natural languages, and it provides various resources for Indian languages like Hindi, Marathi, Bangla. Analysis of multiword expression for the repositories provided by NLTK and n-gram approach was used to carry out the processing of Hindi text and then further for analysis of Multiword Expressions [25]. Sentiment analysis of code mix scripts for Indian languages like Hindi, Marathi is attempted ([26], [27]). A framework was proposed for Hindi sentiment analysis using Hindi SentiWordNet [28]. Senti wordnet is a publicly available lexical resource for opinion mining [29]. Sentiment analysis for Hindi Tweets [30] and Hindi language [31] is explored by researchers. Many individual researchers and institutions have come up with wordnets that have the sentiments tagged. The wordnets distinguish words as neutral, positive and negative. In the songs, the sentiments are more complex. The wordnets must be inclusive of such sentiments to model complex sentiments.

Mood classification of Hindi songs based on lyrics provided polarity of songs in positive or negative sentiments [32]. Similar research work was carried out [33] to analyze the lyrics of Hindi-language-based songs to detect the mood. A mood taxonomy is used to distinguish songs into Happy or Sad. Data is given to the LDA model to discover the hidden emotions within each song. Music mood [34], a system to predict the mood of songs using machine learning, was proposed. It provided sentiment analysis of songs in different eras based on happy or sad moods. TF IDF approach in natural language processing was used for automatic mood classification of lyrics [35] and similar work [36] into four classes based on Russel's model of valence and arousal. A model was proposed for classifying poems in the Hindi language based on navras (nine classes) as per Indian literature [37]. The multimodal approach for sentiment analysis of nursery rhymes using acoustic features and lyrics provided positive and negative sentiments [38]. Researchers explored a multimodal approach using audio features and Hindi lyrics data ([39], [40]).

According to a literature survey done, it was observed that there is little research work done related to Devanagari script text analysis. Therefore, it was necessary to find a wordnet or a collection with words tagged with their sentiments for sentiment analysis. The wordnets available in Hindi have limited capabilities considering the lyrics analysis task. Thus, they are not sufficient enough to tag words or find their meaning for a large dataset. For mood analysis, it is essential to have wordnet where the words are tagged to some meaning or mood. To overcome this issue, individuals have resorted to translation to the English language. Some have resorted to transliteration to avoid the hassle of using the Devanagari script. Though this may solve the issue of wordnet as the English wordnet is vast, it still comes with its problems. Some sentences lose their actual meaning. Some English words mean different things in Hindi when transliterated. These words could cause discrepancies when translated and used for analysis with the help of English wordnet.

The Contextual analysis is done with different approaches, and similar terminologies are explored depending on the task. Standard algorithms and codes for English language processing are readily available, but they are not helpful for the proposed work considering the text data to be processed in the Devanagari script used for Hindi. Moreover, standard datasets for Hindi lyrics in the Devanagari script are not readily available in the public domain. Thus, the need was identified to develop datasets and algorithms for the Devanagari script handling with the Hindi language as a central source of information.

## Dataset

Unlike the English language, there is a dearth in a repository of textual data for Hindi. Even though one might come across data in English script for Hindi songs, Devanagari script data was scarce. Thus, it was necessary to build a dataset for the proposed experiments. We studied different lyrics and tagged the words to the sentiments that belonged to the five moods. These categories represent the song sentiments in a broader sense and not just the negative and positive emotions. Thus, the dataset of a total of 300 songs belonging to 5 moods was created. The dataset is publicly available for further exploration to other researchers [5]. These songs belong to 5 moods, namely happy, sad, party, romantic and devotional. This dataset is further divided into training and testing datasets. The training dataset consists of 250 songs, with 50 songs per mood and the testing with ten songs per mood. The training dataset is used to prepare the knowledge base, which contains words and their probability to occur in a song belonging to one of the five determined moods. The song lyrics in the text files are in the Devanagari script. These files are encoded in the UTF-8 format of character encoding. Figure 1 shows sample data stored in each text file in the Devanagari script.

Contextual mood analysis for Hindi song lyrics in Devanagari scriptचला जाता हूँ, किसी की धुन में
धड़कते दिल के, तराने लिये
मिलन की मस्ती, भरी आँखों में
हज़ारों सपने, सुहाने लिये, चला जाता हूँ ...

ये मस्ती के, नज़ारें हैं, तो ऐसे में
सम्भलना कैसा मेरी क़सम
तू लहराती, डगरिया हो, तो फिर क्यूँ ना
चलूँ मैं बहका बहका रे
मेरे जीवन में, ये शाम आई है
मुहब्बत वाले, ज़माने लिये, चला जाता हूँ ...

वो आलम भी, अजब होगा, वो जब मेरे
करीब आएगी मेरी क़सम
कभी बइयाँ छुड़ा लेगी, कभी हँसके
गले से लग जाएगी हाय
मेरी बाहों में, मचल जाएगी
वो सच्चे झूठे बहाने लिये, चला जाता हूँ ...

बहारों में, नज़ारों में, नज़र डालूँ
तो ऐसा लागे मेरी क़सम
वो नैनों में, भरे काजल, घूँघट खोले
खडी हैं मेरे आगे रे
शरम से बोझल झुकी पलकों में
जवाँ रातों के फ़साने लिये, चला जाता हूँ ...

Fig 1: Example of data in text files

## Methodology

The sentiment context identification of Hindi songs by lyrics is a challenging task. The system should be able to recognize or predict the mood from the lyrics. A three-stage approach, viz Tokenization, Normalization and Feature extraction, was followed to achieve the goal. Text tokenization means to break down sentences into small units such as words (tokens) or phrases (n-grams). The Word-Tokenizer and Regexp Tokenizer functions from nltk libraries in python were used for tokenization. In order to clean the data, punctuations and stop words were identified and removed as they are irrelevant in analyzing the sentiment. Various Stemmers/Lemmatizers are used in NLP to extract the meaning of words such as Porter Stemmer, Snowball Stemmer or Wordnet Lemmatizer. These techniques were not applied to lyrics text as it is likely to miss the inherent meaning and context.

The text files are used to store the lyrics of the songs. These songs were in the Devanagari script; hence the encoding of the files was changed to UTF-8. Finally, the knowledge base is stored in CSV files.

The text files from the training data set were converted into the knowledge base used to conduct sentiment analysis. First, the text data file is broken down into unigrams, i.e., single terms. Then the data is cleaned by removing all the punctuation marks and stop words. This cleaned data is then processed further and added to the .csv file, which will act as our knowledge base. It contains columns like happy, sad, party, devotional, romantic, total, prob_happy, prob_sad,

prob_party, prob_devotional, and prob_romantic. The columns with the name of the genre contain the number of times a word occurs in the lyrics text files of that genre. The columns with the prefix "prob_" followed by the genre name contains the probability of the word occurring in that particular genre and the column "total" number of times the word has occurred in the entire training data set. Different vectorizers are such as Count (Term-Frequency), IF-IDF (Term Frequency- Inverse Document Frequency), TF-IDF + POS (parts of speech) selection are used to identify the weights or importance of words in the specific context. The system uses the TF- IDF approach to associate weights or probabilities to each word.

As an example, if the word 'दुःख' may have appeared five times in the sad songs data set and two times in romantic songs, and the total occurrence of the word is seven times in the entire data set. Thus, the probability of occurrence of the word to associate specific mood is computed with a simple probability formula, as shown in Table 1.

Table 1: Probability occurrence

| word | prob_happy | prob_sad | prob_romantic | prob_devotional | prob_party |
|------|------------|----------|---------------|-----------------|------------|
| दुःख | 0 | 0.714 | 0.285 | 0 | 0 |

The following algorithm is used to create the knowledge base from the lyrics data.

1. *Start.*
2. *Set data <= [NULL]*
3. *If input.txt is not empty*
    a. *data <= Read(input.txt)*
    b. *tokenize(data)*
    c. *clean(data)*
4. *Set freuency_table <= dictionary ()*
5. *i <= 0*
6. *count <= data.count(data[i])*
7. *frequency_table[token] <= count*
8. *i <= i-1*
9. *if i is not equal to length(data)-1 go to step 6*
10. *Set genre_list <= [happy, sad, romantic, party, devotional]*
11. *Set genre <= genre_input.txt and kb <= dataframe (columns = 'word', genre_list, 'Total')*

**Contextual mood analysis for Hindi song lyrics in Devanagari script**

12. kb <= Read(kbase.csv)
13. i= length (frequency_table)
14. if token in kb
    a. Where kb['word'] == token
        i. kb ['genre'] <= kb ['genre'] + frequency_table[token]
15. else
    a. index <= length(kb)
    b. If index not equal to 0
        i. kb [index+1, 'word'] <= token
        ii. kb [index+1, 'genre'] <= frequency_table[token]
    c. else
        i. kb [index, 'word'] <= token
        ii. kb [index, 'genre'] <= frequency_table[token]
16. Set i <= 0
17. genre1 <= genre_list[i]
18. kb['Total'] <= kb ['Total'] + kb[genre1]
19. i <= i + 1
20. if i is not equal to length (genre_list)-1 go to step 17.
21. set k <= 0
22. genre2 <= genre_list[k]
23. prob_gen = concatenate (string ('prob_'), string (genre2))
24. kb [prob_gen] <= kb [genre2] / kb['Total']
25. k <= k + 1
26. if k is not equal to length (genre_list)-1 go to step 22.
27. Stop.

After creating a knowledge base, as shown in Figure 2, it is time to create a model that uses it to determine the sentiment of a song as the input. An unknown text file containing lyrics in the Devanagari script is used as the input to this function. The data in the text file is first broken down into unigrams and then cleaned off all the punctuations. Then the number of occurrences of each term in the input file is calculated. Thus, creating a frequency table containing the word and its number of occurrences. Each term is then searched in the knowledge base. The probability of term occurrence is found out for each mood. If the word is not in the knowledge base, then the word is dropped. Only the words in the knowledge base are further used to predict the sentiment of the song. The frequency of the term is then multiplied by the probability for each genre obtained from the knowledge base. Then the probability of each genre for every term in the song is then

**Contextual mood analysis for Hindi song lyrics in Devanagari script**

summed up. The genre with the highest probability is considered to be the sentiment of the song.

Once the model predicts the genre for an unknown song lyric, lyric data is added to the knowledge base as incremental learning. The frequency table created in the earlier stage of analysis is used for this purpose. If the words exist in the knowledge base, the count of a term in the predicted genre is modified accordingly. If it does not exist, then the word is added as a new record in the list and its probability. The values in prob_happy, prob_sad, prob_party, prob_devotional and prob_romantic columns are recomputed again.

Fig. 2: Screenshot of the knowledge base file

**Contextual mood analysis for Hindi song lyrics in Devanagari script**

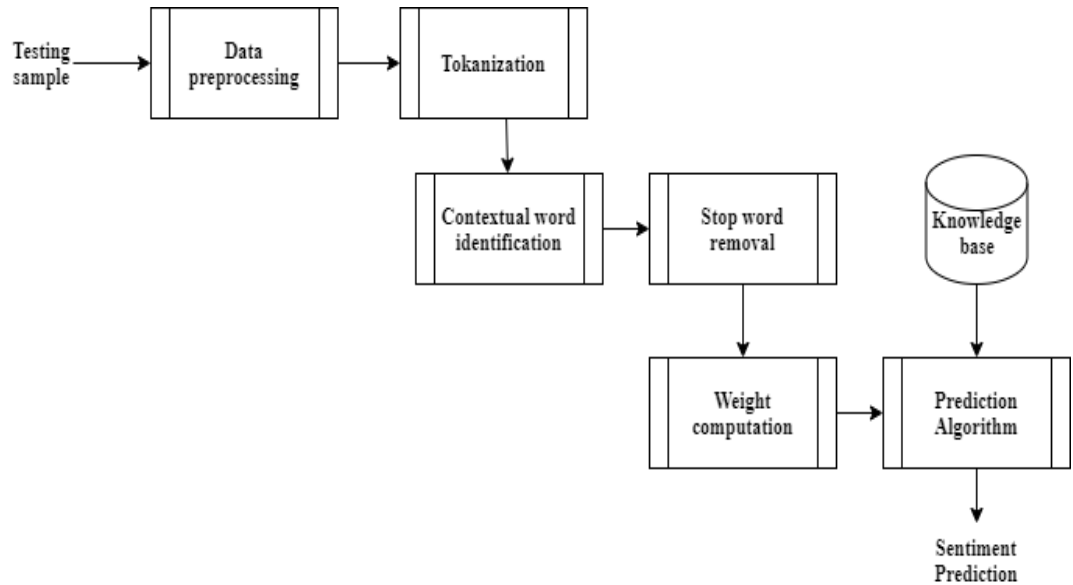

Fig 3: Incremental learning system

The following algorithm illustrates the steps to identify the sentiment of the song.

1. Start.
2. Set data <= [NULL]
3. If input.txt is not empty
   a. data <= Read(input.txt)
   b. tokenize(data)
   c. clean(data)
4. Set freuency_table <= dictionary ()
5. freuency_table <= frequency(data)
6. kb <= Read(kbase.csv)
7. Set i <= 0 and tokens <= [NULL]
8. tokens <= frequency_table.keys()
9. genre_list <= [happy, sad, romantic, party, devotional]
10. Set probability_table <= dataframe(columns = 'word', genre_list)
11. if token[i] in kb
    a. Set k <=0
    b. genre1 <= genre_list[k]
    c. prob_gen = concatenate (string ('prob_'), string (genre1))

    d. *probability_table[token[i], genre1] <=frequency_table[token[i]] *kb[token[i], prob_gen]*

    e. *k <= k +1*

    f. *if k not equal to length(genre_list)-1 go to step b.*

12. *Set sum <= dictionary ()*

13. *Set j <= 0 and i <=0*

14. *sum [genre_list [i]] = probability_table [genre_list[i], j + 1]*

15. *i <= i +1*

16. *if i is not equal to length(genre_list)-1 got to step 14.*

17. *k <= k+1*

18. *if j is not equal to length(probability_table), set i <= 0 and go to step 14.*

19. *Set maximum = max(sum.values())*

20. *Set Sentiment_of_song <=sum.key[maximum]*

21. *Display Sentiment_of_song.*

22. *Stop*

## Result and discussion

The accuracy of the model provides the success of the proposed system. For this, the original 300 songs data set was divided into training and testing data sets. The training data set was used to obtain the knowledge base. The testing data files were then passed to the sentiment analysis function individually to obtain the predicted genres. Then the number of positive results (the results where the predicted genre was the same as the known or the actual genre of the song) was calculated to find the accuracy. When tested for 50 training songs, the accuracy comes out to be 64% for the test data set. The confusion matrix for the test data is as shown in Table 2.

Table 2: Confusion matrix

| **Predicted/ Actual** | Happy | Sad | Romantic | Devotional | Party |
|---|---|---|---|---|---|
| Happy | **4** | 1 | 2 | 0 | 3 |
| Sad | 2 | **4** | 4 | 0 | 0 |
| Romantic | 1 | 1 | **8** | 0 | 0 |
| Devotional | 0 | 0 | 0 | **10** | 0 |

**Contextual mood analysis for Hindi song lyrics in Devanagari script**

| Party | 4 | 0 | 0 | 0 | **6** |
|---|---|---|---|---|---|

Data visualization helps conclude the results better. The challenge is to draw diagrams that show the relationship between lyrics and sentiments. Knowledge Graphs serve this purpose perfectly. Many tools can be used for this purpose, such as python matplotlib. However, the python graphs can be used to display the data but lacked storage and retrieval functionality. Thus, graph databases were explored and used for knowledge representation. A graph database for data visualization purposes would aid in visualization and help store the data, ensuring easy access to the analyzed data. Neo4j is an ACID-compliant transactional database, and it is implemented in java and is open-source software. It uses Cypher query language to query the database, which is similar to SQL. The song's title and mood (sentiment) are used as the nodes in the graph. The genres viz happy, sad, party, devotional and romantic are the nodes to which the title of the songs is connected. The edge (relation) between the song title and the genre exists if the song belongs to that genre (for the songs in the training dataset, one can say that it is known that they belong to that genre). The terms which depict those sentiments are also connected to the title of the song.

The knowledge graph tool used is Neo4j [41]. The specialty of a graph database is that data can be stored in a graph and retrieved using a query language. Thus, the time used to analyze a known song from the training set can be reduced by passing a search query to the graph database. Furthermore, the answer obtained from the search query can be displayed as the analysis and provide faster results. The python API for neo4j is used to interact with the knowledge graph. The queries can be passed using the API, and the results obtained can be printed onto the console, as shown in figure 4.

**Contextual mood analysis for Hindi song lyrics in Devanagari script**

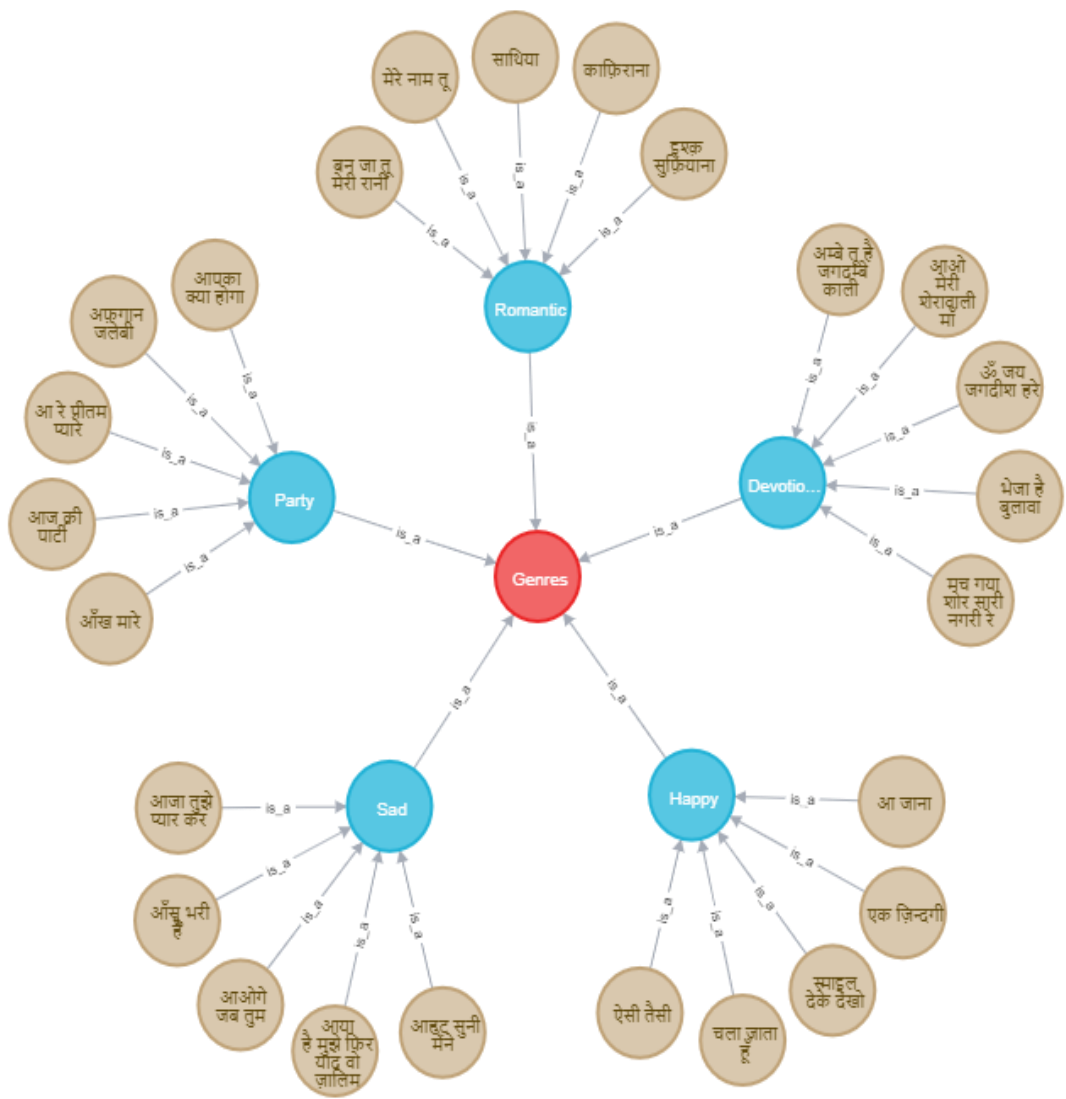

Fig 4: Knowledge graph

# Conclusion and future directions

The work intends to contribute to the research in the Hindi language processing in Devanagari script. The data set and proposed algorithms contribute to this field. The analysis model and its accuracy validate the proposed system. The work can be further extended to enhance accuracy. The dataset can be used by other researchers who wish to explore and work on the Devanagari script. The knowledge graph can be used as a senti wordnet. The words are tagged to the sentiments. Thus, it can be used to analyze more complex sentiments apart from the positive and negative ones. More sentiments too can be added to this list, and it can be increased to more categories.

This textual analysis can be clubbed with audio analysis of music to get better accuracy in determining the song's mood. It can be used in the entertainment sector for making playlists according to mood or taste. Though the current work focuses on the lyrics data, this knowledge can be applied to any Hindi textual data. Be it tweets, news, or any Hindi literature work can be accessed to find out the theme of the textual data. It can help to categorize Hindi data according to their context.

# Contextual mood analysis for Hindi song lyrics in Devanagari script